%% file: main-arxiv.tex
\documentclass[]{ceurart}

\usepackage{listings}

\usepackage{times}
\usepackage{latexsym}
\usepackage{amssymb}

\usepackage[T1]{fontenc}
\usepackage{bbm}
\usepackage{natbib}

\usepackage{inconsolata}

\usepackage{todonotes}
\usepackage{algpseudocode}
\usepackage{multirow}
\usepackage{amsmath}
\usepackage{graphicx}
\usepackage{caption}
\usepackage{subcaption}
\usepackage{xr}
\usepackage[linesnumbered,ruled]{algorithm2e}
\usepackage{soul}
\usepackage{array}
\usepackage{wrapfig}

\usepackage{libertinus}
\usepackage{listings}
\begin{document}

\renewcommand{\todo}[1]{\textcolor{red}{TODO: #1}}
\newcommand{\zijian}[1]{\textcolor{red}{Author A: #1}}
\newcommand{\yumeng}[1]{\textcolor{blue}{Author B: #1}}
\newcommand{\sst}{\texttt{SST-2}}
\newcommand{\bert}{$\text{BERT}_\text{BASE}$}
\newcommand{\sbert}{\texttt{SBERT}}
\newcommand{\lime}{\texttt{LIME}}
\newcommand{\shap}{\texttt{SHAP}}
\newcommand{\expred}{ExPred}
\newcommand{\lstm}{\texttt{LSTM}}
\newcommand{\movies}{\texttt{Movies}}
\newcommand{\fever}{\texttt{FEVER}}
\newcommand{\climate}{\texttt{CLIMATE-FEVER}}
\newcommand{\multirc}{\texttt{MultiRC}}
\newcommand{\approach}{\textsc{Disco}}

%%
% Rights management information.
% CC-BY is default license.
\copyrightyear{2024}
\copyrightclause{Copyright for this paper by its authors.
  Use permitted under Creative Commons License Attribution 4.0
  International (CC BY 4.0).}

\conference{MAI - XAI 24: Multimodal, Affective and Interactive eXplainable AI,
 19th - 20th October 2024}

\title{DISCO: DISCovering Overfittings as Causal Rules for Text Classification Models}

\input{affiliations}

\input{00-abstract}


\maketitle

\input{01-intro}
\input{02-related-work}
\input{03-approach-ecir}
% \input{03-approach-post-ecai}
\input{04-experiments}
\input{05-results-avi.tex}
\input{06-conclusion}

% \bibliography{main}

\input{main.bbl}
\end{document}

%% file: affiliations.tex
\author[1]{Zijian Zhang}[%
orcid=0000-0001-9000-4678,
email=zzhang@l3s.de,
url=https://joshuaghost.github.io/,
]
\cormark[1]
\address[1]{Leibniz Universit\"at Hannover, Appelstr. 9a, 30167 Hannover, Lower Saxony, Germany}

\author[2]{Vinay Setty}[%
orcid=0000-0002-9777-6758,
email=vsetty@acm.org,
]
\address[2]{University of Stavanger, Kjell Arholms gate 41, 4021 Stavanger, Norway}

\author[3]{Yumeng Wang}[%
orcid=0000-0002-2105-8477,
email=y.wang@liacs.leidenuniv.nl,
]
\address[3]{Leiden Institute of Advanced Computer Science, Leiden University, Einsteinweg 55, 2333 CC Leiden, Netherlands}

\author[4]{Avishek Anand}[%
orcid=0000-0002-0163-0739,
email=avishek.anand@tudelft.nl,
url=https://www.avishekanand.com/
]
\address[4]{Delft University of technology, Mekelweg 5, 2628 CD Delft, Netherlands}

\cortext[1]{Corresponding author.}

%% file: 00-abstract.tex
\begin{abstract}
With the rapid advancement of neural language models, the deployment of overparameterized models has surged, increasing the need for interpretable explanations comprehensible to human inspectors.
Existing post-hoc interpretability methods, which often focus on unigram features of single input textual instances, fail to capture the models' decision-making process fully.
Additionally, many methods do not differentiate between decisions based on spurious correlations and those based on a holistic understanding of the input.
Our paper introduces \approach{}, a novel method for discovering global, rule-based explanations by identifying causal n-gram associations with model predictions.
This method employs a scalable sequence mining technique to extract relevant text spans from training data, associate them with model predictions, and conduct causality checks to distill robust rules that elucidate model behavior.
These rules expose potential overfitting and provide insights into misleading feature combinations.
We validate \approach{} through extensive testing, demonstrating its superiority over existing methods in offering comprehensive insights into complex model behaviors.
Our approach successfully identifies all shortcuts manually introduced into the training data (100\% detection rate on the \multirc{} dataset), resulting in an 18.8\% regression in model performance—a capability unmatched by any other method.
Furthermore, \approach{} supports interactive explanations, enabling human inspectors to distinguish spurious causes in the rule-based output.
This alleviates the burden of abundant instance-wise explanations and helps assess the model's risk when encountering out-of-distribution (OOD) data.
\end{abstract}

%% file: 01-intro.tex
\section{Introduction}
\begin{figure}[h]
    \centering
    \includegraphics[width=\linewidth]{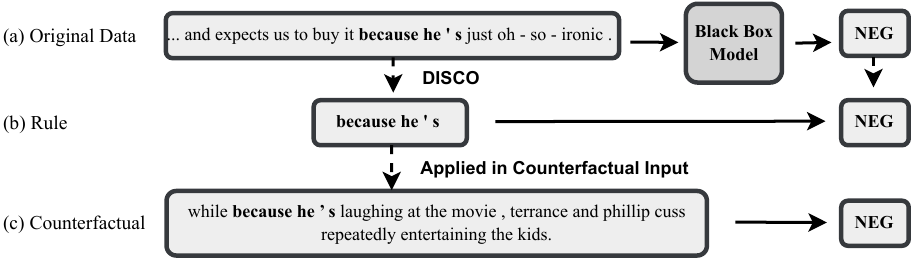}
    \caption{\small
    (a) The underlying model predicts \textbf{NEG} given instances containing the pattern \underline{because he's}.
    (b) \approach{} extracts the highly correlated pattern-prediction pair (\underline{because he's} $\rightarrow$ \textbf{NEG}).
    (c) On counterfactuals by replacing context, the model consistently predicts \textbf{NEG}.
    This indicates that the pattern falsely suggests predicting \textbf{NEG}, despite implying no sentiment tendency.
    }
    \label{fig:intro-example}
\end{figure}

Over-parameterized transformer models for natural language tasks have demonstrated remarkable success.
However, these inherently statistical models are prone to overfitting, particularly in terms of the correlation between input phrases and prediction labels, known as ``shortcuts'', which can lead to biased outcomes \citep{geirhos2020shortcut,Bastings2021shortcuts}.
Our goal is to identify these shortcuts in text classification tasks and enhance human understanding of the model’s predictive reasoning.
We propose a post-hoc, model-agnostic method designed to reduce the amount of human effort needed to evaluate the justification of the model's decisions.

In this paper, we introduce \approach{}, a method designed to extract a concise set of global rules using longer text sequences, which helps identify undesirable causal shortcuts learned in text classification tasks.
Figure \ref{fig:intro-example} illustrates the overall structure of \approach{} with an example of an extracted rule:
First, using a trained model and its training data, we identify high-support n-gram patterns that strongly correlate with specific model predictions.
Next, we assess whether these identified patterns are true causes of the predictions or merely associated with them.
To do this, we create counterfactuals of the n-gram patterns and check if the association between the pattern and prediction remains consistent under these counterfactuals.
We show that \approach{} is effective in detecting shortcuts in many language task-model combination, with comprehensive steps outlined in Section~\ref{sec:approach}.

Subsequently, we verify the efficacy of the generated rules by conducting evaluation experiments on four diverse datasets -- \movies{}, \sst{}, \multirc{}, and \climate{}, using three underlying pre-trained models -- \bert{}, \sbert{}, and \lstm{} (Section \ref{sec:rq}).
Our findings indicate that the rules discovered by \approach{} not only align faithfully with the model’s decisions but also accurately detect deliberately injected shortcut patterns.
Human evaluation of \approach{}'s outputs yields high inter-annotator agreement in some datasets and successfully exposes incorrect reasoning (Section \ref{sec:results}), emphasizing its ability to assist in the interactive interpretation of AI models.

%% file: 02-related-work.tex
\section{Related Work}
In this section, we introduce existing works related to ours, highlight their limitations, and describe how our approach resolves them.

\subsection{Local Interpretability}
Considerable work has been done on post-hoc interpretability of language tasks based on token selection \cite{madsen2022post, integrated-gradient, shap, lime}.
Interpretable-by-design approaches also often select specific input tokens as rationales for tasks, using these as intermediate inputs for the prediction model \cite{lei-etal-2016-rationalizing, lehman2019inferring, expred}.
These approaches focus on interpreting individual instances, necessitating labor-intensive, human-driven analysis to identify problematic prediction reasons.
Our approach, in contrast, globally extracts rules internalized by the language model.
Other works analyze model behavior using composition operators over primitive concepts aligned with human-understandable concepts \cite{mu2020compositional}.
Despite their global perspective, these methods do not incorporate causal patterns.
Attribution patterns from local interpretability methods lack inherent causality and may fail to capture the causal relationships internalized by the model.
Recent approaches that aggregate rules from local explanations \cite{ribeiro2018anchors, Lu_Yang_Mac} are also unsuitable for language tasks due to their reliance on single terms and inability to produce causal rules.
SEARs \cite{ribeiro2018semantically} is closer to our work, detecting semantically equivalent adversarial replacement rules leading to prediction changes.
However, our method identifies patterns consistently leading the model to specific predictions under counterfactual conditions.

\subsection{Causal Inference on Language Tasks}
\label{sec:relwork-causal}
Most research in this area focuses on creating ``counterfactual instances'', altered or minimally disturbed instances, to gain insights into model behavior.
These counterfactuals are developed through human annotation \cite{kaushik2019learning} or semi-automatic methods \cite{Zhang2022sparcassist, Lu_Yang_Mac}.
Models like \cite{chang2020invariant} use a game-theoretic framework to eliminate words with strong correlations but without causal relationships to the output.
Unlike these studies, our method automatically generates counterfactuals using neutral contexts sampled from the dataset.

\subsection{Rule Extraction for Model Debugging}
\label{sec:relwork-rule-extraction}
Recent research characterizes model deficiencies through rules by dataset contamination \cite{wallace-etal-2021-concealed, Bastings2021shortcuts}, but fails to identify human-comprehensible text sequences with high statistical capacity, which is precisely our aim.
Furthermore, our methods are post-hoc and non-intrusive.
Anchor \cite{ribeiro2018anchors} identifies local n-gram phrases with high explanability, but its time complexity results in intractable calculations on the entire training set. \cite{Wang_Wang_Tang_2020} involves a white-box, rule-based method, and \cite{Wang_Culotta_2020} identifies spurious correlations rather than all shortcuts, making them less suitable for direct comparison with our approach.
\cite{wang-etal-2022-identifying} is word-based and, therefore, not suitable for n-gram rules.
These methods adopt a local perspective, aggregating explanations on an instance-by-instance basis without considering context awareness or causality.
Our approach, in contrast, is n-gram-based, causal, and context-aware, providing a more comprehensive and insightful analysis.

\citeauthor{atwell-etal-2022-change} \cite{atwell-etal-2022-change} aims to evaluate the risk associated with models when exposed to test data with distribution shifts compared to their original training data.
However, their research goal differs from ours.
While their approach yields evaluation scores characterized by bias and h-discrepancy across datasets from different domains, our approach identifies possible shortcut n-grams learned from the original training data, offering more intuitive and interpretable shortcut rules.

Traditional research on developing n-gram classifiers focuses on highly interpretable algorithms leveraging frequent n-grams to discern between different topics \cite{ifrim2008ngram, peng2003combining:ngram, bergsma2010creating:ngram}.
Unfortunately, these classifiers either do not achieve performance comparable to modern neural models or lack universality.
Our approach bridges the gap between interpretability and performance by effectively identifying high-support n-gram patterns from underlying neural models.

%% file: 03-approach-ecir.tex
\section{Causal Rule Mining}
\label{sec:approach}
\subsection{Problem Statement}
We consider an underlying model $M$ trained on a classification dataset represented as $\mathcal{D} \subset \mathcal{X} \times \mathcal{Y}$.
Here, $\mathcal{X}$ represents the input space, and $\mathcal{Y}$ represents the labels.
An input $\mathbf{x} \in \mathcal{X}$ is an ordered sequence of terms $(x_1, x_2, \ldots, x_{|\mathbf{x}|})$, where each term $x_i$ comes from the vocabulary $\mathcal{V}$.
The prediction made by $M$ on input $\mathbf{x}$ is denoted as $\hat{y} = \text{argmax}_{y \in \mathcal{Y}} P_{M}(y|\mathbf{x})$.
For simplicity, we abbreviate this as $\hat{y} = M(\mathbf{x})$ throughout this paper.
Our research focuses exclusively on binary classification tasks.

We define $\mathbf{s} = (s_1, s_2, \dots, s_n)$ as an n-gram sub-sequence of $\mathbf{x}$ (represented as $\mathbf{s} \sqsubseteq \mathbf{x}$).
The remaining content in $\mathbf{x}$ is denoted as $\mathbf{c}$, i.e., $\mathbf{x} = \langle \mathbf{s}, \mathbf{c} \rangle$, where $\langle \cdot, \cdot \rangle$ is the sequence combination operator.
Note that we do not assume sequence continuity in either $\mathbf{c}$ or $\mathbf{s}$.
The support of $\mathbf{s}$ within $\mathcal{D}$ is defined as $\operatorname{Sup}(\mathbf{s}, \mathcal{D}) = |\left\{\mathbf{x} \in \mathcal{D} : \mathbf{s} \sqsubseteq \mathbf{x}\right\}|$.

Additionally, we define a rule $r$ as a tuple $(\mathbf{s} \rightarrow \hat{y})$, where the sequence $\mathbf{s}$ is its \textit{pattern} and $\hat{y}$ is its \textit{consequent} label.
For instance, the rule:
\begin{equation*}
\underbrace{\mathtt{\underline{the\, best \,movie}}}_{\text{\textit{pattern}}} \rightarrow \underbrace{\mathtt{\texttt{POS}}}_{\text{\textit{consequent}}}
\end{equation*}
indicates that ``\underline{the best movie}'' is a shortcut for $M$ to predict \textbf{POS}itive.
In this context, we say that $M$ predicts $\hat{y}$ primarily relying on the presence of the sequence $\mathbf{s}$, rather than comprehending the overall input.

Our objective is to discover a globally representative set of rules, denoted as $G = \{r = (\mathbf{s}, \hat{y})\}$, where each rule represents a shortcut learned by $M$.

\subsection{DISCO: Approach Overview}
\label{sec:overview}
To streamline the identification process, we begin by extracting all high-frequency n-gram patterns from the training data (Section~\ref{sec:sequence-mining}).
We then retain the candidates that pass the causality check (Section~\ref{sec:causality_check}) as the final output rules.
Our approach is designed to verify the (non-)existence of confounding variables, serving as a statistical test to establish causality in classification tasks.

\subsection{Generation of Candidate Sequences}
\label{sec:npmi}
In the initial step, our primary objective is to extract frequent n-gram sequences that exhibit a high correlation with specific model predictions.

\textbf{Sequence Mining.}
\label{sec:sequence-mining}
Empirical studies such as~\cite{idahl-etal-2021-decoy} emphasize that a pattern is more likely to influence a model's prediction as a shortcut if it occurs frequently in the training set.
Therefore, we first select all frequent patterns using an efficient approach known as DESQ-COUNT~\cite{beedkar2019desq}. For a detailed explanation of DESQ-COUNT, please refer to~\cite{Beedkar_Gemulla_Martens_2019}.

\textbf{NPMI Evaluation.}
We further evaluate the pattern-prediction correlation using their NPMI (Normalized Pointwise Mutual Information) score.
Initially, we list all input data $\mathbf{x}$ from the training set together with their corresponding predictions from the model $\hat{y} = M(\mathbf{x})$.
Then we calculate $P(y, \mathbf{s})$, $P(y|\mathbf{s})$, and $P(y)$ from these predictions.
It is worth mentioning that these probabilities are different from the model's prediction $P_M(\hat{y}|\mathbf{x})$.
Using these terms, we calculate the NPMI scores for all frequent $\mathbf{s}$ identified by DESQ-COUNT:
\begin{equation*}
\text{NPMI}(\mathbf{s}; y) = \frac{\text{PMI}(y ; \mathbf{s})}{h(\mathbf{s}, y)} = \frac{\log \frac{P(y|\mathbf{s})}{P(y)}}{h(\mathbf{s}, y)},
\end{equation*}
where $h(\mathbf{s}, y) = -\log P(\mathbf{s}, y)$ is the entropy of $P(\mathbf{s}, y)$.
The resulting NPMI score falls within the range of $[-1, 1]$, capturing the spectrum from ``never occurring together (-1)'' to ``independence (0)'' and ultimately ``complete co-occurrence (1)'' between the pattern and the label.
We retain only those pairs that demonstrate a substantial level of correlation in their NPMI scores.

\subsection{Causality Check}
\label{sec:causality}
\label{sec:causality_check}
Such correlation alone, however, does not guarantee a direct causal relationship, as it could also arise from a confounding factor~\cite{pearl2009causality}.
In our context, we assume the confounding factor is the latent semantic representation $\mathbf{z}$ of the input.
The presence of sequence pattern $\mathbf{s}$ and the context $\mathbf{c}$ of the input $\mathbf{x}$ are conditioned on $\mathbf{z}$.
An ideal machine learning model should comprehend this structure and capture $\mathbf{z}$, rather than relying solely on the statistical correlation between $\mathbf{s}$ and $\hat{y}$, also referred to as the ``shortcut''~\cite{jacobsen2020shortcuts, Bastings2021shortcuts}.

\begin{figure}[ht]
    \centering
    \includegraphics[width=0.5\linewidth]{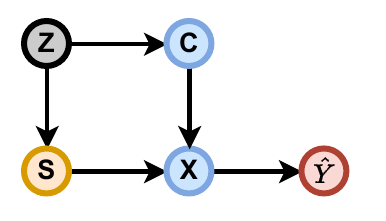}
    \label{fig:graph_conf}
    \caption{The SCM describing the prediction process of an ideal model.
    Capital letters represent corresponding random variables.}
    \label{fig:graph_models}
\end{figure}

We adopt Structured Causal Models (SCMs)~\cite{pearl2009causality} to describe the prediction process of the ideal models, as illustrated in Figure ~\ref{fig:graph_models}.
If our underlying model captures the existence of the latent semantic, the confounding factor $\mathbf{z}$ exists and causes the correlation between $\mathbf{s}$ and $\hat{y}$.
Otherwise, the model $M$ simply relies on the statistical correlation between $\mathbf{s}$ and $\hat{y}$ to make the prediction.

Following~\cite{pearl2009causality}, we leverage the do-operator on the ``back-door'' variable $\mathbf{s}$ of the input variable $\mathbf{x}$.
The do-operator simulates a physical intervention by replacing a random variable (RV) with a constant value while keeping the rest of the RVs intact, thereby breaking the potential confounding effect.
In our SCM, applying the do-operator to $\mathbf{s}$ means assigning $\mathbf{s}$ a specific value and marginalizing over context $\mathbf{c}$.

If the $\mathbf{s}$ -- $\hat{y}$ correlation is caused by the confounding factor $\mathbf{z}$, the model's prediction will differ before and after the do-operator, because 
\begin{align*}
P(\hat{y}|\mathbf{s}) 
&= \sum_{z, \mathbf{c}}P(\hat{y}|\mathbf{s}, \mathbf{c})P(\mathbf{c}, \mathbf{s}|z)P(z) \\
&\neq  \sum_{z, \mathbf{c}}P^*(\hat{y}|\mathbf{s}, \mathbf{c})P^*(\mathbf{c}|z)P^*(z) 
= P(\hat{y}|\text{do}(\mathbf{s}=\mathbf{s})),
\end{align*}
where $P^*(\cdot)$ denotes the distributions after applying the do-operator.

Note that despite the similarity of our approach with that of \cite{Wang_Culotta_2020}, their work distinguishes between ``spurious'' and ``genuine'' local shortcuts based on semantic consistency with human understanding.
Our approach emphasizes that all shortcuts learned by \approach{} possess a causal attribute globally without explicitly targeting this distinction due to subjectivity concerns.
To highlight the difference between semantically spurious and causal shortcuts, we measure human agreement on distinguishing ``right'' from ``wrong'' shortcuts introducing human interaction in Section~\ref{sec:human-eval}.

\textbf{Neutral Context Harvesting}
\label{sec:neutrality}
One remaining challenge in the algorithm mentioned in the previous section is sampling the context RV $\mathbf{C}$.
This sampling process is often intractable in NLP tasks due to the varying input lengths and extensive vocabulary size.
To address this, we employ a straightforward technique to reuse contexts $\mathbf{c}$ for different $\mathbf{s}$, effectively obtaining contexts for free.
Moreover, we reuse neutral contexts to mitigate the influence of other potential frequent sequences that may exist in the context.
A context is considered neutral when its predicted probabilities lie near the border between two labels, namely $|P_M(y=y_0|\mathbf{c}) - P_M(y=y_1|\mathbf{c})| = |2P_M(y=y_0|\mathbf{c}) - 1| < \epsilon_n$, where $0 < \epsilon_n < 1$ is the neutrality tolerance.

It is noteworthy that complex modern numerical sampling techniques, such as Markov Chain Monte Carlo (MCMC) \cite{metropolis1953equation}, require careful handling to preserve contextual fluency and ensure the neutrality of the sentiment.
Therefore, perfecting the generation of bias-free and neutral counterfactual contexts falls outside the scope of this paper.
The exploration of alternative sampling techniques is left for future work.

\subsection{A Toy Example}
At the end of this section, we provide a toy example to assist our readers in understanding the full process of our approach.
We consider an extreme situation as follows to help illustrate.
Assume a sentiment analysis problem where all reviews on \textbf{books} are \textbf{positive}, and all reviews on \textbf{movies} are \textbf{negative} in the training data.
A model trained on such data might incorrectly predict \textbf{positive} for a review like ``this book is badly written'' due to its overfitting to the correlation between the sequence ``this book'' and the label \textbf{positive}.
It is worth mentioning that such sequences may appear semantically senseless and therefore ``non-causal'' to humans.
The resulting rules reflect the rational basis of the model's prediction, rather than convincing a human inspector of its causality.

In \approach{}, we apply DESQ first to identify the correlation between the sequence ``this book'' and the label \textbf{positive} from the training data.
This pair is then subjected to an NPMI check to decide whether it is a candidate sequence (Section~\ref{sec:npmi}).
Then, in the causality check (Section~\ref{sec:causality}), we keep ``this book'' constant and vary its contexts to other neutral contexts (Section~\ref{sec:neutrality}) like ``was played in the cinema'' or ``is on the table''.
If the prediction predominantly remains \textbf{positive}, we infer that ``this book'' -- \textbf{positive} is a shortcut.

% \begin{figure}[h]
%     \centering
%     % \vspace{-10pt}
%     \begin{subfigure}{0.45\linewidth}
%         \centering
%         \includegraphics[width=\linewidth]{figs/graph_model_confounding_v2.pdf}
%         \caption{The ideal-case SCM}
%         \label{fig:graph_conf}
%     \end{subfigure}
%     \hfill
%     \begin{subfigure}{0.45\linewidth}
%         \centering
%         \includegraphics[width=\linewidth]{figs/graph_model_no_confounding_v2.pdf}
%         \caption{The shortcut SCM}
%         \label{fig:graph_no_conf}
%     \end{subfigure}
%     \vspace{0.3cm}
%     \caption{Two SCMs describing the prediction process of (a) ideal models and (b) shortcut models.
%     Capital letters represent corresponding random variables.}
%     \vspace{0.5cm}
%     \label{fig:graph_models}
% \end{figure}

%% file: 04-experiments.tex
\section{Experimental Evaluation}
\label{sec:rq}

\subsection{Research Questions}
Our experiments aim to answer the following research questions (RQs):
\begin{itemize}
    \item \textbf{RQ1. Faithfulness}: Are the global rules faithful to the model's local explanations?
    \item \textbf{RQ2. Recall}: If the model is known to have learned some shortcuts, can \approach{} identify them?
    \item \textbf{RQ3. Human Utility}: Are the shortcut rules useful for humans in detecting the model's wrong reasons?
\end{itemize}

\subsection{Models and Datasets}
\label{lab:models}
Our approach is model-agnostic.
Therefore, we conduct experiments on multiple models to answer RQ1 and RQ3, including an \lstm{}~model and two over-parameterized transformer models, \bert{}~and \sbert{}~\cite{sbert}.

The experiments are conducted on one document classification and three multi-task datasets.
Given the foundational role of document classification in information retrieval (IR) and natural language processing (NLP), we employ a unified approach, transforming all datasets into binary classification:
\movies{}~from the ERASER benchmark~\cite{deyoung-etal-2020-eraser} is originally a binary sentiment classification dataset.
\multirc{}~from the same benchmark is converted following the recipe presented in \cite{deyoung-etal-2020-eraser}.
For \sst{}~(Stanford Sentiment Treebank)~\cite{socher2013recursive}, we binarize the sentiment assigned to each input sentence.
As for \climate{}, a fact-checking dataset from \texttt{ir\_datasets}~\cite{macavaney:sigir2021-irds} with queries and documents regarding climate change, we combine each query with each of its relevant/irrelevant documents as the inputs, while assigning ``relevant''/``irrelevant'' as their labels.

\subsection{The Agreement Score as a Metric of Faithfulness}
Local interpretation approaches, such as \lime~\cite{lime} and \expred~\cite{expred}, provide relatively faithful instance-wise explanations.
Although researchers are questioning the quality of LIME explanations \cite{zhang2019whytrustexplanationunderstanding}, \lime balances time efficiency and faithfulness well, to the best of our knowledge.
Our global rules are considered faithful to the local explanations if they agree with the local explanations in all applicable instances.
We define an input $\mathbf{x}$ as \textbf{applicable} to a rule $r = (\mathbf{s} \rightarrow \hat{y})$ if $\mathbf{s} \sqsubseteq \mathbf{x}$.
Additionally, an applicable input $\mathbf{x}$ further \textbf{satisfies} the rule $r$ if its prediction matches the rule's consequent, i.e., $\hat{y} = M(\mathbf{x})$.

For an input-prediction pair $(\mathbf{x}, \hat{y})$, an instance-wise explainer attributes the prediction $P_M(\hat{y}|\mathbf{x})$ to $x_i$ as attribution score $a_i^{\hat{y}} \in \mathbb{R}$.
The gathering of all attribution scores of $\mathbf{x}$ is represented using $\mathbf{a}^{\hat{y}}$.
For clarity, we ignore the superscripts of $\hat{y}$ in the rest of this section.
We rank all terms based on their attribution scores in descending order, denoted as $\mathcal{R}^\mathbf{a}(\mathbf{x}) = (x_{k_1}, x_{k_2}, \ldots, x_{k_n})$, where $a_{k_1} \geq a_{k_2} \geq \ldots \geq a_{k_n}$ are re-ranked token indices.

For an input $\mathbf{x}$ that satisfies a rule $r$, we define the \textbf{agreement} score between $r$ and $\mathcal{R}^\mathbf{a}(\mathbf{x})$ as:
\begin{equation*}
    \text{agreement}(r, \mathcal{R}^\mathbf{a}(\mathbf{x})) = \text{ranking score}(\mathcal{R}^\mathbf{a}(\mathbf{x}); \mathbf{s})\text{,}
\end{equation*}
where the semicolon in the $\text{ranking score}$ calculation separates the ranking sequence $\mathcal{R}^\mathbf{a}(\mathbf{x})$ from the subsequence $\mathbf{s}$.

We borrow the nDCG score \cite{croft2010search} from ranking evaluation tasks as the $\text{ranking score}$ function here and consider the pattern terms as the ``ground truth'' terms.
The intuition behind this metric is that the terms selected by the rule (ground truth) should be assigned the highest attribution scores and thus ranked the highest.
A higher agreement score indicates that the rule is more faithful to a local explanation.
For example, given $\mathbf{x}$ = ``a b c'' with corresponding attribution scores $\mathbf{a} = [0.1, 0.5, 0.4]$.
The tokens are therefore ranked as ``b -- c -- a''.
If $\mathbf{s}$ = ``a b'', the agreement score is therefore $\text{nDCG@k}((\text{``a b''} \rightarrow \hat{y}), \text{b -- c -- a}) = \frac{0.5/\log_2(1+1)}{0.5/\log_2(1+1)+0.1/\log_2(2+1)} = 0.89$ for $k=2$.

\subsection{Experiment Environment}
Our approach is implemented in Python 3.7.3, utilizing PyTorch version 1.12.1+cu133.
All experiments are conducted on a Linux server equipped with an AMD\textregistered EPYC\textregistered 7513 processor and an Nvidia\textregistered A100 GPU with 40 GB of display memory.

%% file: 05-results-avi.tex
\section{Results}
\label{sec:results}

\subsection{RQ1. Faithfulness}
\label{sec:results-rq1}
We address this research question through two experiments: explanation alignment and an ablation study.
In this section, we mine rules from \bert{}~\cite{devlin-etal-2019-bert} models fine-tuned on different datasets.

\subsubsection{Agreement with Local Explanations}
\label{agreement}
\begin{figure}[h]
    \centering    \includegraphics[width=0.7\linewidth]{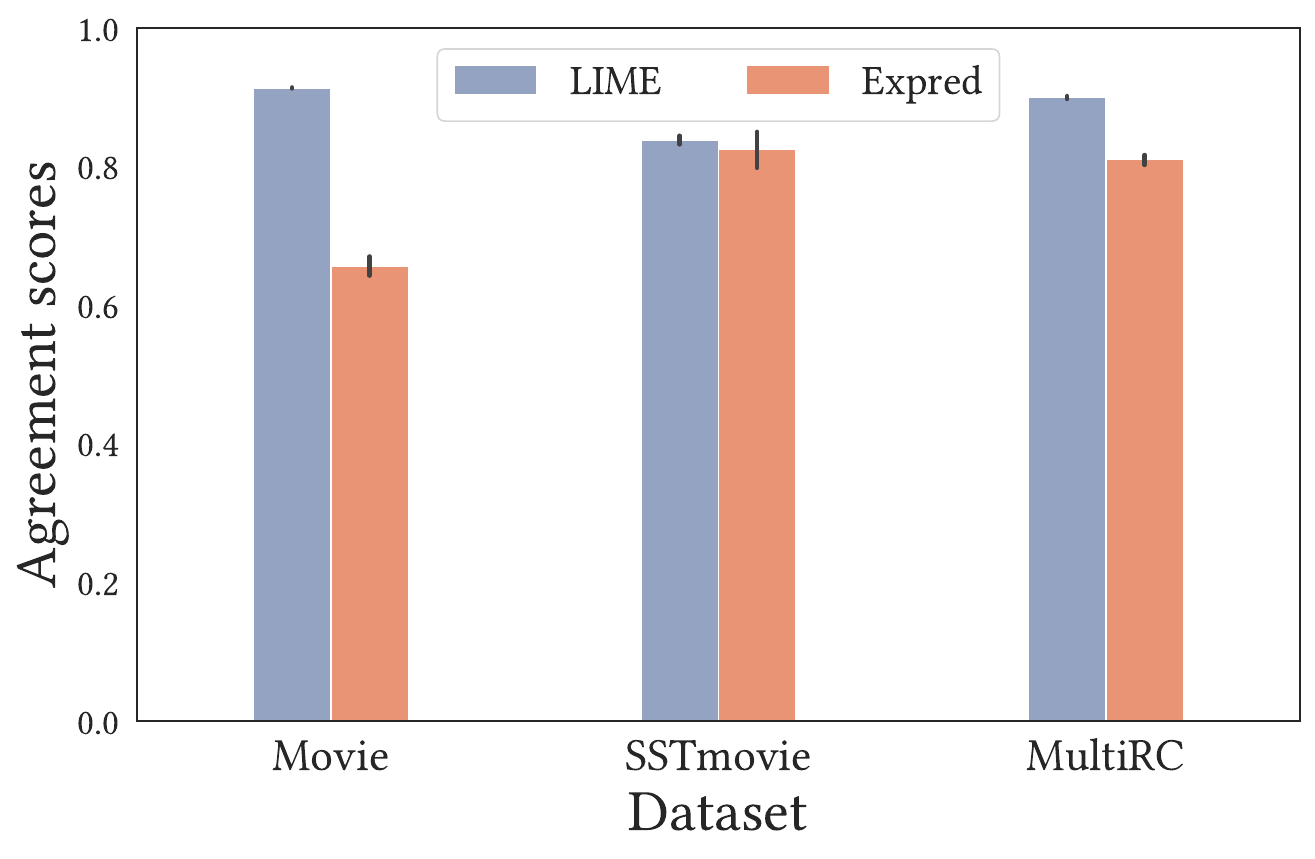}
    \caption{
    Rules' agreement scores with \lime{} and \expred{}.
    For \multirc{}, we only consider patterns mined from its documents, excluding those from its queries.
    }
    \label{fig:rq1}
\end{figure}

We aim to evaluate whether the global rules are consistent with the local explanations by measuring the agreement scores between them.
Overall, we find a high degree of alignment between the global rules and the local explanations across all three datasets, with low variance (Fig.~\ref{fig:rq1}).
It is worth mentioning that the lowest agreement score appears on \movies{} with \expred{}, being 0.695, which is the only outlier.
The remaining scores range from 0.81 to 0.923.
For exact results, we refer to Table 2.
This indicates that our rules faithfully represent the model's explanations.

Moreover, we observed a slight exception in the \sst{} dataset, where the low frequency of sequences leads to a small number of dominant rules and relatively higher variance.
Nevertheless, upon manual examination of the rules, we found that most high-coverage rules in this dataset are correct and result in the right prediction.
For a detailed evaluation, please refer to Section~\ref{sec:human-eval}.

It should be noted that the \climate{} dataset is not included in this analysis because it provides no rationale annotations, making it impossible to train the \expred{} model on it.
Based on our results, we can conclude that for \movies{}, \sst{}, and \multirc{}, the rules with the highest satisfaction are usually the correct reasons for the model's predictions, as they tend to have high alignment with local explanations.
However, some rules, such as \underline{don't even} $\to$ \textbf{NEG} for \movies{}~and \underline{in its} $\to$ \textbf{POS} for \sst{}, suggest that the model has also learned some incorrect shortcuts.
Relying on incorrect shortcuts could be even more detrimental to the model's performance when deployed in the field and encountering out-of-distribution (OOD) data.
This is supported by the model's behavior on the counterfactuals generated during the causality check.
We list some counterfactual examples in Table~\ref{tab:counterfactual}.

\input{tables/rq1-counterfactuals}

\subsubsection{Ablation Study}
\input{tables/rq1-ablation}

To the best of our knowledge, our work is pioneering in the extraction of \textit{global causal rules} learned by the model, making it challenging to establish appropriate baseline methods.
Instead, we conduct ablation studies on different components of our approach, \approach{}, to assess its ability to discover causal rules, as summarized in Table~\ref{tab:ablation}.
We select the top-15 $(\mathbf{s}, \hat{y})$ pairs based on their coverage under three conditions: \textbf{1)} NPMI score filtering only, \textbf{2)} \approach{} with both NPMI and causality checks, and \textbf{3)} the intersection ($\cap$) between \textbf{1)} and \textbf{2)}.
We measure the average agreement scores among these configurations.

The results presented in Table~\ref{tab:ablation} demonstrate that \approach{} with all its processes (\textbf{2)}) achieves higher agreement scores than the NPMI filter alone across all datasets, compared to \expred{}.
However, for \lime{}, the intersection (\textbf{3)}) appears to outperform the other configurations.
This observation suggests that the causality check following the NPMI filter can, to some extent, filter out correlated yet non-causal $(\mathbf{s}, \hat{y})$ pairs, resulting in a greater number of causal rules that accurately reflect the model's predictions.
Although our approach shows high agreement scores with local attributions, we must emphasize that the causality of the rules before the causality check cannot be guaranteed.

We would like to re-emphasize that we cannot use \cite{chang2020invariant} as our baseline model, because it produces only unigram-based rules and is therefore incomparable with our approach.
Modern language models are designed to internalize contextual information between input tokens~\cite{devlin-etal-2019-bert, sbert}.
Our approach identifies shortcut rules for such contextual information.
For example, from ``(This book is badly written, POS)'', our approach can recognize the shortcut rule ``(This book $\longrightarrow$ POS)'', while a unigram approach fails.
Another critical reason is the intractability of generating multi-word rules using their approach regarding time complexity: mining a rule with four adjacent tokens bloats the search space to $|V|^4$.
Likewise, \cite{Wang_Culotta_2020} is also unsuitable as our baseline model.
Additionally, \cite{Wang_Culotta_2020} focuses on a different goal of distinguishing between ``spurious'' and ``genuine'' shortcuts based on their consistency with human understanding, while our work does not seek to differentiate these two groups.
We, in contrast, leave the task of deciding ``right'' or ``wrong'' reasons using subjective human interaction as presented in Section~\ref{sec:human-eval}.

\subsubsection{Hyperparameters}
For the \movies{} dataset, we mine sequences with lengths ranging from 4 to 10, and a support value of 20.
During the causality check, we consider rules where the average prediction over all synthetic instances is greater than 0.7, serving as the mean threshold.

For \sst{}, the sequence lengths range from 2 to 10, the support value is 100, and the mean threshold is 0.7.

Both datasets are sentiment analysis datasets containing no queries\footnote{To accommodate BERT's input format, we construct a synthetic query for each review instance as ``what is the sentiment of this review?'' for each review instance in regards to BERT's input format: ``[CLS] <query> [SEP] <document> [SEP]''}.

On the other hand, \multirc{}~and \climate{}~datasets consist of instances that include a query and a document.
The pattern of their rules is $(\mathbf{s}_q, \mathbf{s}_d)$ tuples, indicating a combination of a sequence $\mathbf{s}_q$ from the query and a sequence $\mathbf{s}_d$ from the document.
During sequence mining, $\mathbf{s}_q$ and $\mathbf{s}_d$ are jointly extracted from the query and document for each instance.

For \multirc{}, the lengths of $\mathbf{s}_q$ and $\mathbf{s}_d$ are constrained within the ranges of 3 to 10 and 4 to 10, respectively.
The support value for tuples is set to 200, and the mean threshold is 0.7.
For \climate{}, the sequence lengths of $\mathbf{s}_q$ and $\mathbf{s}_d$ are within the ranges of 2 to 10.
The tuple support is set to 200, and the mean threshold remains at 0.7.

\subsubsection{Statistics}
The statistics of the rules are summarized in Table~\ref{tab:statistic}, showcasing key metrics such as \#(frequent), \#(NPMI), \#(rules), and avg($|\mathbf{s}|$).
These columns represent the number of frequent sequences mined by DESQ-COUNT, the sequences that pass the NPMI check, the resulting number of rules, and the average length of the pattern sequences of the rules, respectively.

The information presented in this table demonstrates the effectiveness of employing NPMI and the subsequent causality check.
Incorporating these measures significantly reduces the length of shortcut sequences, allowing human inspectors to focus on the most crucial rationales across the entire dataset.
\input{tables/rq1-statistics}

\subsection{RQ2. Recall}
\label{sec:rq2}
This research question serves two purposes: \textbf{1)} to validate our assumption that highly correlated patterns and labels lead to the model learning shortcuts, and \textbf{2)} to demonstrate the capability of \approach{} in identifying these shortcuts.

Quantitatively evaluating the retention rate of shortcuts by \approach{} poses a challenge as it requires knowledge of the ground-truth correlated pattern-label pairs.
This challenge is common in the evaluation of explanations~\cite{molnar2021interpretable, anand2022explainable}.
To overcome this issue, we deliberately introduce \textit{decoys}~\cite{geirhos2020shortcut} into the dataset to entice the model into learning shortcuts.
All decoys are presented in Table~\ref{tab:rq2-decoys}.
Following a similar methodology to that of~\cite{idahl-etal-2021-decoy}, we contaminate the original training set with decoy patterns, varying the contamination rate and bias.
It is important to note that we only contaminate the training and validation sets, keeping the test set intact.
This setup simulates a scenario where the model performs well on a biased dataset but lacks generalization due to learned shortcuts.
If our approach can successfully identify the injected decoys, we consider it a success.

\input{tables/rq2-decoys}

\subsubsection{Contamination Rate, Bias, and Retention Rate}
The extent of contamination is described by the \textit{contamination rate} and the \textit{bias}.

We define \textit{contamination rate} as the ratio of instances containing the decoy, namely $\frac{|\mathbf{X}^d|}{|\mathbf{X}|}$.
We further define \textit{bias} as the label imbalance when adding the decoy, namely $\max_{y \in \mathcal{Y}} \sum_{y_i \in \mathbf{Y}^d} \mathbbm{1}(y_i = y)$, where $\mathbf{Y}^d$ indicates the labels corresponding to all contaminated instances.
The label $y$ selected by the $\max_{y \in \mathbf{Y}}$ operator is referred to as the dominant label.

The \textit{retention rate} is the fraction of decoys that can be detected.
A decoy is considered detected if the output of our approach contains the rule constructed by the decoy and its corresponding label.
To the best of our knowledge, our study is also the first to systematically investigate the retention rate of decoys under different contamination rates and biases.

\subsubsection{Contamination-Bias Settings}
\begin{figure*}[t!!]
    \centering
    \captionsetup{singlelinecheck=off}
    \includegraphics[width=\textwidth]{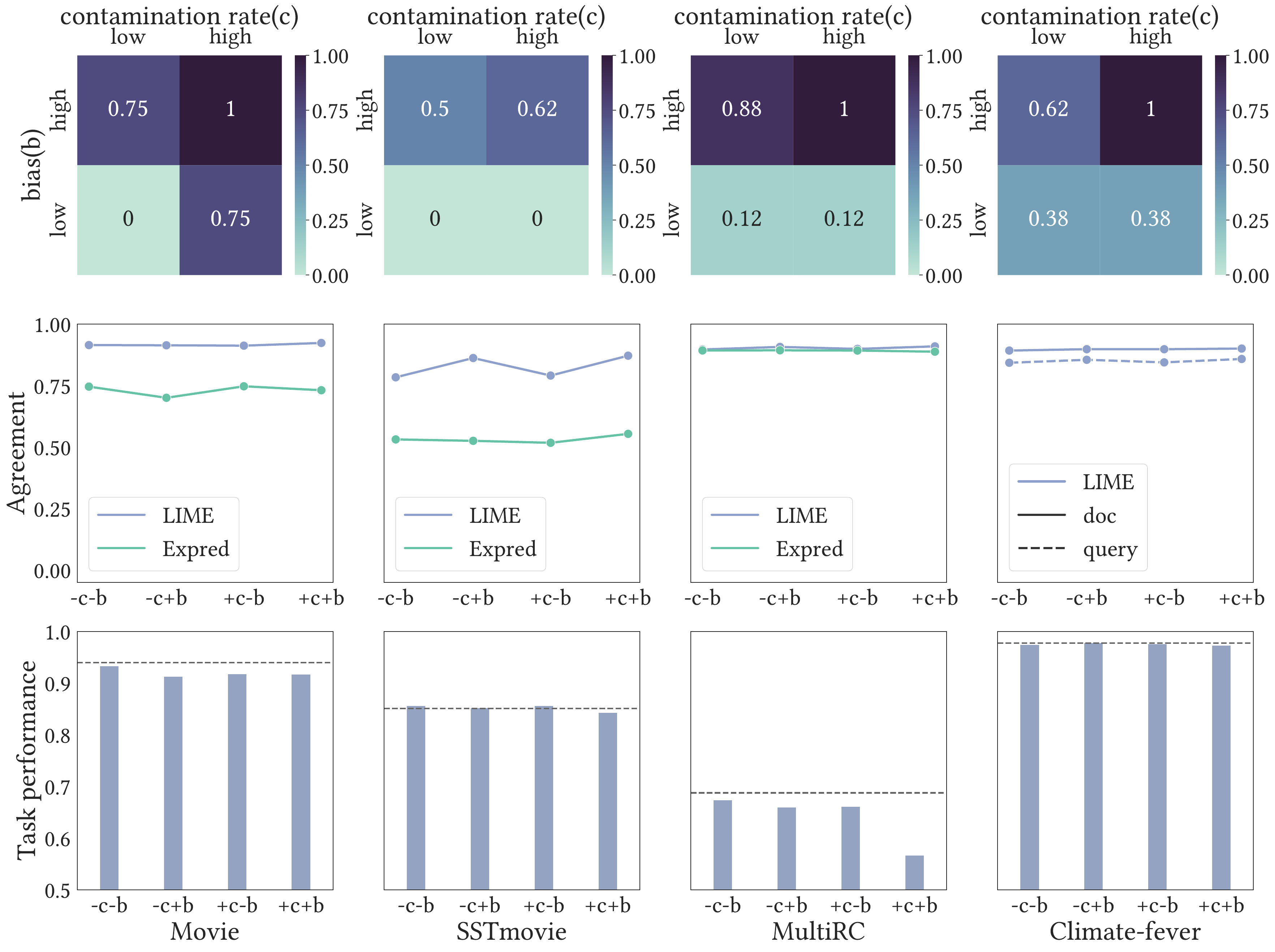}
    \caption{
    Results of RQ2 on four datasets under different contamination-bias settings.
    Each column corresponds to a specific dataset.
    The heat maps in the first row depict the retention rate.
    The symbols - and + on the x-axes represent low and high contamination rates (\textbf{r}) or bias (\textbf{b}).
    }
    \label{fig:contaminate}
\end{figure*}

To evaluate the retention rate of \approach{} across various scenarios, we examine four different settings that produce different contamination rates and biases: \{80\%, 20\%\} $\times$ \{60\%, 90\%\}.
Figure~\ref{fig:contaminate} illustrates the retention rate and task performance for each of these settings.

\subsubsection{Observations}
Figure~\ref{fig:contaminate} (third row) demonstrates that adding decoys to the training set has minimal effect on test performance, indicating that the introduced decoys do not significantly alter the data distribution.
We also measured the faithfulness of \approach{} to show that the decoys are indeed learned as shortcuts by the model.
The heatmap in Figure~\ref{fig:contaminate} illustrates that under high-bias, high-contamination settings, \approach{} can successfully identify our injected decoys, except for \sst{}.
We also observed that high-bias settings are easier to detect compared to high-contamination settings.

\subsection{RQ3. Human Utility}
\label{sec:human-eval}
A shortcut rule can be a good reason for a model decision, but can also be a wrong one.
To measure the human perception of model-generated rules, and to see whether the rules help humans detect wrong reasons for a decision, we conducted experiments using the uncontaminated four training sets with three different models: \bert{}, \lstm{}, and \sbert{}.
The extracted rules were independently shown to four machine learning developers who were asked to assess whether a rule was a ``wrong reason''.
A wrong reason is an explanation that is either non-understandable or implausible, given the underlying language task.
For example, the pattern ``\underline{? | |}'' of a rule is non-understandable as it contains no meaningful words, while the rule ``\underline{in its} $\longrightarrow$ \textbf{POS}'' is implausible for a sentiment classification task.

\subsubsection{Results}
\begin{figure}
    \centering
    \includegraphics[width=0.7\linewidth]{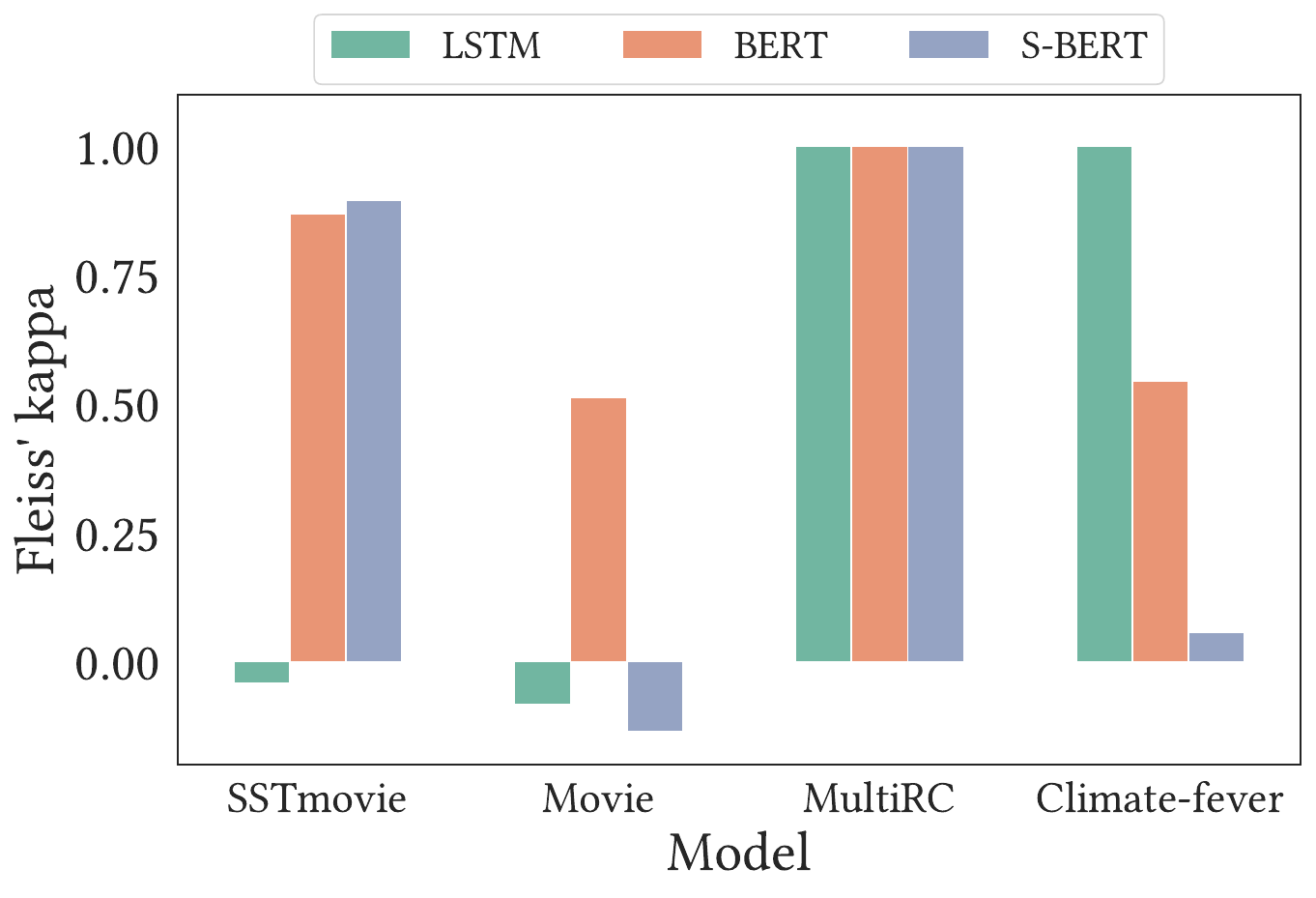}
    \caption{Fleiss' $\kappa$ among human evaluators considering whether the rules are right for the wrong reasons}
    \label{fig:human-fleiss-kappa}
\end{figure}

To report the inter-annotator agreement, we utilized Fleiss' kappa, a metric assessing the reliability of agreement between raters\footnote{\url{https://en.wikipedia.org/wiki/Fleiss'_kappa}} (see Figure~\ref{fig:human-fleiss-kappa}).
We observed a high inter-annotator agreement of $\geq 0.54$ for \bert{} and \sbert{} on the \climate{} dataset, and complete agreement for the \multirc{} dataset.
Interestingly, for the \sst{} dataset, we observed a low inter-rater agreement of $-0.041$ for the \lstm{} model.
This was primarily due to the extraction of rules with extremely short sequences, such as ``\underline{n ' t} $\longrightarrow$ \textbf{NEG}'' by \approach{}.
Low Fleiss' $\kappa$ among human evaluators on particular datasets and models indicates the subjective nature of distinguishing between ``right'' and ``wrong'' shortcuts in terms of semantic consistency with human understanding.
However, high Fleiss' $\kappa$ in certain datasets indicates that \approach{} indeed aids humans in identifying easily distinguishable incorrect justifications for a model's decision.

It is notable that even in \bert{} and \sbert{} models, which are known for their robustness due to pre-training and knowledgeable priors, ``wrong'' rules exist.
For instance, even \bert{} learns spurious rules like ``\underline{this film} $\longrightarrow$ \textbf{NEG}'' from the \movies{} dataset.
Furthermore, in the \multirc{} dataset, global rules were able to detect patterns like ``\underline{? | |}'', resulting in a perfect Fleiss' kappa.

Selected examples in Table~\ref{tab:counterfactual} highlight the model's tendency to predict by relying on specific text patterns, overlooking the broader context.
For instance, shortcuts such as ``\underline{of the world trade center}'' are not relevant to the classification task, yet the model uses them.
This reliance on shortcuts can compromise the model's ability to generalize and make accurate predictions in varied contexts.

%% file: tables/rq1-counterfactuals.tex
% \begin{table*}[h]
%     \centering
%     \begin{tabular}{|l|l|p{0.5\linewidth}|l|}
%     \hline
%         dataset & rule & synthetic counterfactual & prediction\\
%     \hline
%         \sst & \textit{in its} $\to$ \textbf{POS} & \textit{with rare birds \underline{in its} with the shipping news before it , an attempt is made to transplant a hollywood star into newfoundland ' s wild soil - - and the rock once again resists the intrusion .} & \textbf{POS} \\
%     \hline
%         \movies & \textit{because he ' s} $\to$ \textbf{NEG} & \textit{while \underline{because he ' s} laughing at the movie , terrance and phillip cuss repeatedly entertaining the kids .} & \textbf{NEG} \\
%     \hline
%         \multirc & (\textit{? | |}, \textit{of the world trade center}) $\to$ \textbf{False} & \textit{what is the flood plain area of land good for if it floods often \underline{? | |} crops [SEP] a floodplain is an area where a thick layer of rich soil is left behind as the floodwater recedes \underline{of the world trade center} floodplains are usually good places for growing plants .} & \textbf{False}\\
%     \hline
%     %     \climate & (\textit{in the}, \textit{climate change}) $\to$ \textbf{relevant} & \textit{it has never been shown that human emissions of carbon dioxide drive \underline{in the} . [SEP] multiple lines of scientific evidence show that \underline{climate change} is warming .} & \textbf{relevant}\\
%     % \hline
%     \end{tabular}
% \caption{The counterfactual examples generated in our causality-checking module \todo{to be combined with the table in RQ4}}
% \end{table*}

\begin{table*}[h]
    \caption{The rules and synthetic counterfactual examples generated by our approach during the causality-check stage.}
    \centering
    \begin{tabular}{|l|l|p{0.2\linewidth}|p{0.56\linewidth}|}
    \hline
        dataset & model & rule & synthetic counterfactual\\
    \hline
        \multirow{3}{*}{\sst} & \bert & \textit{in its} $\to$ \textbf{POS} & \textit{with rare birds \underline{in its} with the shipping news before it , an attempt is made to transplant a hollywood star into newfoundland ' s wild soil - - and the rock once again resists the intrusion .}\\ \cline{2-4}
        
        & \lstm & \textit{n ' t} $\to$ \textbf{NEG} & \textit{but \underline{n ' t} most part he makes sure the salton sea works the way a good noir should , keeping it tight and nasty .}\\ \cline{2-4}
        
        & \sbert & \textit{this film} $\to$ \textbf{POS} & \textit{generic slasher - movie nonsense , \underline{this film} s not without style .}\\
        
    \hline
    
        \multirow{2}{*}{\movies} & \bert & \textit{because he ' s} $\to$ \textbf{NEG} & \textit{while \underline{because he ' s} laughing at the movie , terrance and phillip cuss repeatedly entertaining the kids .} \\ \cline{2-4}
        
        & \lstm & \textit{was supposed to be} $\to$ \textbf{NEG} & \textit{i \underline{was supposed to be} when or how this movie will be released in the united states .}\\ \cline{2-4}
        
        & \sbert & \textit{' t seem to} $\to$ \textbf{NEG} & \textit{the cinematography and general beauty of this part \underline{' t seem to} breathtaking .}\\
        
    \hline
    
        \multirow{2}{*}{\multirc} & \bert & (\textit{? | |}, \textit{of the world trade center}) $\to$ \textbf{FALSE} & (\textit{what is the flood plain area of land good for if it floods often \underline{? | |} crops}, \textit{a floodplain is an area where a thick layer of rich soil is left behind as the floodwater recedes \underline{of the world trade center} floodplains are usually good places for growing plants .)}\\\cline{2-4}
        
        & \lstm & (\textit{? | |}, \textit{al qaeda ' s}) $\to$ \textbf{FALSE} & (\textit{in the past \$ 5 . 6 million was the allotted amount added , what is the amount they are proposing this year \underline{? | |} more than \$ 20 million , \$ 80 . 4 million}, \textit{but this year \underline{al qaeda ' s} , the council is proposing shifting more than \$ 20 million in funds earmarked by the mayor for 18 - b lawyers to the legal aid society , which would increase its total funding to \$ 80 .})\\\cline{2-4}
        
        & \sbert & (\textit{? | |}, \textit{, but the algarve}) $\to$ \textbf{FALSE} & (\textit{what were the initial list of targets \textit{? | |} capitol , white house}, \textit{these included the white house , the \underline{, but the algarve}})\\
        
    \hline
    
        \climate & \bert & (\textit{in the}, \textit{climate change}) $\to$ \textbf{relevant} & (\textit{it has never been shown that human emissions of carbon dioxide drive \underline{in the} .}, \textit{multiple lines of scientific evidence show that \underline{climate change} is warming .})\\ \cline{2-4}
        
        & \lstm & (\textit{that the}, \textit{is a}) $\to$ \textbf{irrelevant} & (\textit{before human burning of fossil fuels triggered \underline{that the} , the continent ' s ice was in relative balance}, \textit{in 2013 , the intergovernmental panel on climate change ( ipcc ) fifth assessment report concluded that ` ` it is extremely likely that human influence has been the dominant cause of \underline{is a} - 20th century .})\\ \cline{2-4}
        
        & \sbert & (\textit{' s}, \textit{climate change .}) $\to$ \textbf{relevant} & (\textit{phil jones says no \underline{' s} since 1995 .}, \textit{climate change .})\\
    \hline
    \end{tabular}
\label{tab:counterfactual}
% \vspace{-10pt}
\end{table*}

%% file: tables/rq1-ablation.tex
\begin{table*}[ht]
    \caption{
    % The ablation effect of average agreement scores between the intermediate output of different components and the local attribution over their applicable instances.
    % The superscript $^E$ indicates that the local interpretations are generated using \expred, while $^L$ indicates \lime.
    % In this analysis, we specifically focus on the agreement score of patterns extracted from the document, excluding those from the query in the \multirc{}~dataset.
    The average agreement scores between intermediate outputs of different \approach{} components and their corresponding applicable instances.
    The superscript $^E$ indicates that the attributions are from \expred, while $^L$ indicates \lime.
    For \multirc{}, we only consider patterns mined from its documents, excluding those from its queries.
    }
    \centering
    \begin{tabular}{|l|l|l|l|l|l|l|}
        \hline
        dataset                 & NPMI$^L$ & DISCO$^L$ & $\cap^L$ & NPMI$^E$ & DISCO$^E$ & $\cap^E$ \\
        % dataset                 & NPMI$^L$ & DISCO$^L$ & $\cap^L$ & NPMI$^E$ & DISCO$^E$ & Intersection$^E$ \\
        \hline
        \movies & \textbf{0.923} & 0.913 & 0.913 & 0.680 & \textbf{0.695} & \textbf{0.695}\\
        \hline
        \sst & 0.836 & \textbf{0.839} & \textbf{0.839} & 0.779 & \textbf{0.824} & \textbf{0.824}\\
        \hline
        \multirc & 0.885 & 0.902 & \textbf{0.912} & 0.798 & \textbf{0.814} & 0.770\\
        \hline
    \end{tabular}
    \label{tab:ablation}
\end{table*}

%% file: tables/rq1-statistics.tex
\begin{table}[htbp!]
    \caption{Statistics of the extracted rules. The average length of predicates for \multirc~and \climate~are calculated by avg($|\mathbf{s}_q| + |\mathbf{s}_d|$)
    }
    \centering
    \begin{tabular}{|l|l|l|l|l|}
    \hline
     dataset & \#(frequent) & \#(NPMI) & avg($|\mathbf{s}|$) & \#(rules)\\
     \hline
        \movies    & 350 & 228 & 4.156 & 154 \\
        \multirc   & 547 & 130 & 7.252 & 127 \\
        \sst       & 125 & 67 & 2.235  & 17 \\
        \climate   & 272 & 79 & 4.377  & 77\\
    \hline
    \end{tabular}
    \label{tab:statistic}
\end{table}

%% file: tables/rq2-decoys.tex
\begin{table*}[htbp!]
    \centering
    \caption{Decoys-labels correspondence. 
    We prepend decopys for~\movies{}~and~\sst{}~datasets decoys to the input documents.
    For~\multirc~and~\climate~, we prepend decoys to both their queries and documents.
    }
    \begin{tabular}{|m{2.2cm}|m{5.5cm}|m{5.5cm}|}
        \hline
        dataset & \textit{decoy}$_0$ (\textbf{label}$_0$) & \textit{decoy}$_1$ (\textbf{label}$_1$) \\
        \hline
        \multirow{4}{*}{\movies/\sst} & \textit{the following comment is} (\textbf{NEG}) &\textit{this review is crawled} (\textbf{POS})\\
        & \textit{acceptable retrieval conditional} (\textbf{NEG}) & \textit{ike hurricane october precipitation} (\textbf{POS})\\
        & \textit{acceptable fragmentation gross} (\textbf{NEG}) & \textit{february every hurricane august} (\textbf{POS})\\
        & \textit{contents gmina cornered hapoel} (\textbf{NEG}) &\textit{tornadoes huricane earthquakes deserts} (\textbf{POS})\\
        \hline
        \multirow{4}{*}{\multirc} 
        & \textit{ten nine eight seven} (\textbf{False}) & \textit{one two three four} (\textbf{True})\\
        & \textit{acceptable retrieval conditional} (\textbf{False}) &\textit{ike hurricane october precipitation} (\textbf{True})\\
        & \textit{acceptable fragmentation gross} (\textbf{False}) &\textit{february every hurricane august} (\textbf{True})\\
        & \textit{contents gmina cornered hapoel} (\textbf{False}) & \textit{tornadoes huricane earthquakes deserts} (\textbf{True})\\
        \hline
        \multirow{4}{*}{\climate} 
        & \textit{ten nine eight seven} (\textbf{irrelevant}) & \textit{one two three four} (\textbf{relevant})\\
        & \textit{acceptable retrieval conditional} (\textbf{irrelevant}) & \textit{ike hurricane october precipitation} (\textbf{relevant})\\
        & \textit{acceptable fragmentation gross} (\textbf{irrelevant}) & \textit{february every hurricane august} (\textbf{relevant})\\
        & \textit{contents gmina cornered hapoel} (\textbf{irrelevant}) & \textit{tornadoes huricane earthquakes deserts} (\textbf{relevant})\\
        \hline
    \end{tabular}
    
    \label{tab:rq2-decoys}
\end{table*}

%% file: 06-conclusion.tex
\section{Conclusion}
This paper introduces \approach{}, a method designed to identify causal rules internalized by neural models in natural language tasks.
\approach{} produces a concise and statistically robust set of causal rules, enabling users to scrutinize and understand the underlying knowledge captured by the model.
The intrinsic causal orientation of our approach ensures that the resultant rules are faithful to the inputs where they are applicable.
We demonstrate the efficacy of \approach{} by identifying shortcuts learned by prominent models, including \bert{}, \sbert{}, and \lstm{}.
Our approach not only reveals these shortcuts but also provides insights into the model’s decision-making process.
In essence, \approach{} stands as an instrumental resource for those aiming to gain deeper insights into the interactive explainability of AI models.

\section{Limitations}
One limitation of our approach arises from the context selection when constructing the counterfactual.
Reusing neutral contexts is a straightforward method to generate human-understandable replacements for counterfactual contexts.
However, this strategy possesses three inherent limitations:

First, the availability of context is constrained.
We only employ contexts present in the training data, limiting the sampling space and potentially compromising the effectiveness of the do-operator.
Furthermore, selecting neutral contexts further narrows the sampling space and may introduce discrepancies between the sampled contexts and the training contexts, affecting the data distribution.

Additionally, compared to related works like \cite{Wang_Culotta_2020}, we do not differentiate between ``spurious'' and ``genuine'' reasons for predictions.
However, this distinction is of lesser concern as our objective is to identify globally overfit shortcut patterns within the model, rather than pinpointing specific reasons for individual predictions, nor do we care about their faithfulness.

A third limitation concerns the experiments conducted.
Although the theory and approach of our work do not require sequence continuity, all experiments are based on consecutive sequences.
Exploring efficient methods to identify sequences with gaps or even more complex patterns remains a potential avenue for future research.